\documentclass{article}

\usepackage{spconf,amsmath,graphicx}
\usepackage{cite}
\usepackage{algorithmic}
\usepackage{graphicx}
\usepackage{textcomp}
\usepackage{xcolor}
\usepackage{hyperref}
\usepackage{booktabs}
\usepackage{stfloats}
\usepackage{multirow}
\usepackage{amssymb} 

\title{Sparse Attention to Emotion: Efficient Facial Emotion Recognition via Token Reduction}
\name{
Aya Manel Zitouni$^{\ddag}$ \quad
Aicha Zenakhri$^{\ddag}$\footnotemark[1] \quad
Karim Haroun$^{\ddag}$\footnotemark[1]\quad
Larbi Boubchir
\thanks{$^{\ddag}$ Equal contribution.}
}

\address{
LIASD Laboratory, University of Paris 8, France
}

\begin{document}
%
\maketitle

\begin{abstract}
Facial Emotion Recognition (FER) is an important task that has significant implications across various fields such as biometrics, health, and human-computer interaction. Current Vision Transformer-based approaches display quadratic complexity $\mathcal{O}(N^2)$, with N being the input sequence length, making them cumbersome to deploy at the edge. In this paper, we hypothesize that the FER task does not necessarily require all facial information to correctly interpret emotional states, as specific regions such as the eyes, the mouth, and parts of the cheeks carry discriminative information that can be sufficient to recognize emotions. Based on this, we propose Sparse Attention to Emotion (SAE), a model that discards image tokens that have no added value to the emotional context, while preserving good accuracy and achieving a significant gain in computational cost. Surprisingly, even after suppressing 90\% of the image tokens, our model achieves competitive accuracy to state of the art methods at much lower cost, providing a lightweight Facial Emotion Recognition approach. Experimental results demonstrate that SAE achieves new state of the art results on the RAF-DB dataset while reducing the computational complexity by up to 90\%.
\end{abstract}

\begin{keywords}
Vision Transformers, Model Compression, Token Pruning, Facial Emotion Recognition 
\end{keywords}

\section{Introduction}

Facial Emotion Recognition (FER) has received great attention in recent years in various domains. Translating human expressiveness into emotions is a fundamental step in several applications, from personalized experiences to psychological analysis \cite{Ruan_2021_CVPR}, highlighting the importance of the research conducted. Early FER approaches relied on handcrafted features for facial expression analysis. Although effective in controlled settings, such methods often lacked generalization and robustness in real world scenarios \cite{mao2023postersimplerstrongerfacial}. With the advent of deep learning, Convolutional Neural Networks (CNNs) were introduced to improve FER performance by automatically learning discriminative representations \cite{savchenko2021facial}. However, CNN based methods remain limited in capturing long range dependencies due to their inherently local receptive fields \cite{vaswani2017attention, wang2018non}.

To explicitly model the global contextual information of facial expressions, Vision Transformers (ViTs) \cite{dosovitskiy2021an}, which have shown state of the art performance in image recognition, were the inspiration of the proposed Transformer-based FER models \cite{Xue_2021_ICCV,Zheng_2023_ICCV,Her_2025} by processing facial images as sequences of tokens. Among them, POSTER \cite{Zheng_2023_ICCV} achieves state of the art performance by integrating facial landmarks and image features through a two-stream pyramid cross-fusion architecture.

Despite these advances, the large number of image tokens treated indiscriminately to classify emotions, resulting in high computational costs, is a major limitation that remains unaddressed. In this work, we address this problem while maintaining high accuracy at a lower computational cost compared to state of the art methods.

\begin{figure}[!t]
    \centering
    \includegraphics[width=\linewidth]{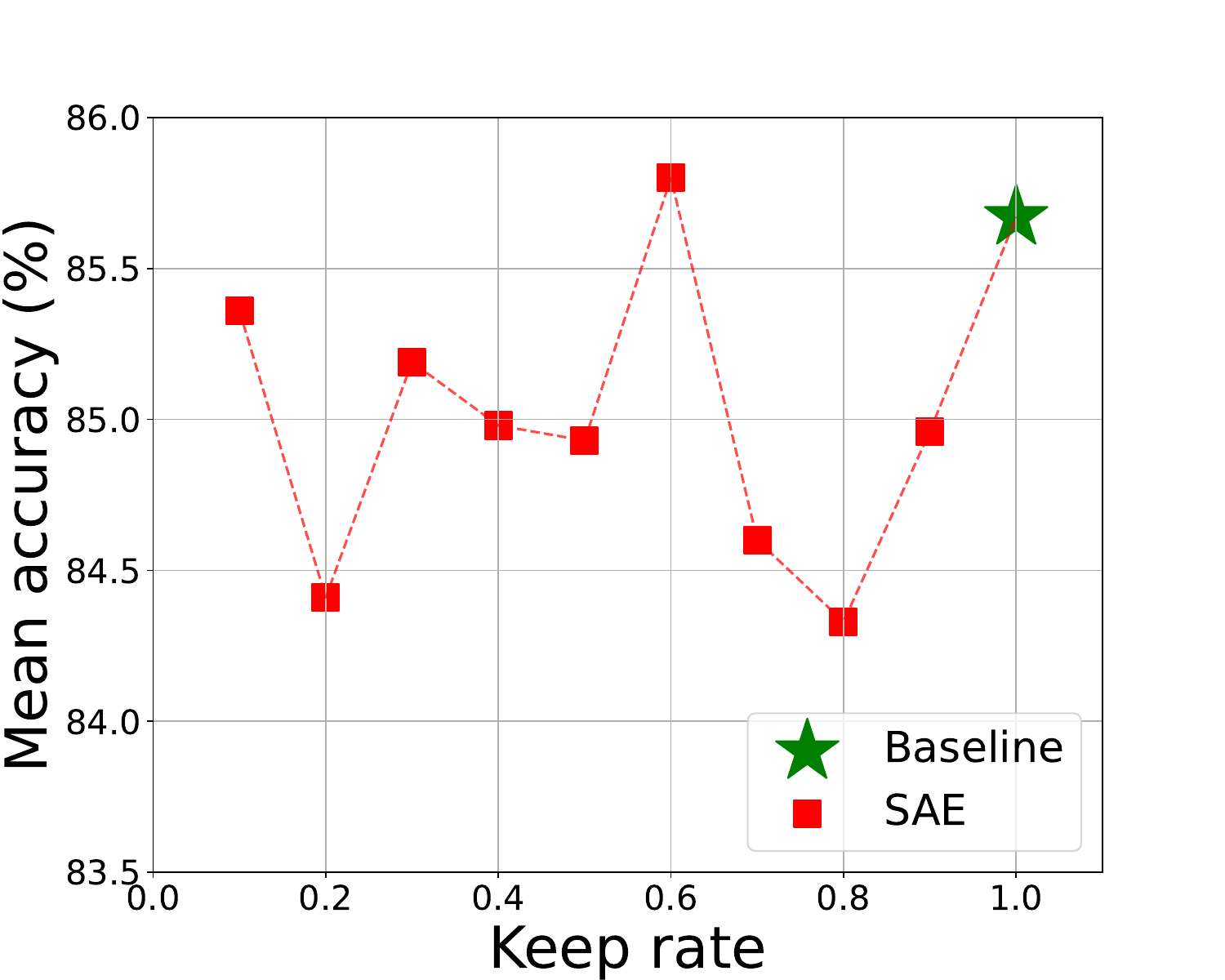}
    \caption{Accuracy vs. keep rate of SAE on Raf-DB dataset. }
    \label{fig:accuracy_kr}
\end{figure}

We propose a token pruning method, originally used for generic image classification, that retains only the tokens with the highest impact on emotion recognition for the FER task and evaluates its performance on standard FER datasets. After experimenting with token pruning at different rates and analyzing the remaining tokens. We observe that in most cases the retained tokens correspond to specific regions of the face. As shown in Figure \ref{fig:accuracy_kr}, when retaining only 10\% of the image tokens, which cover mainly the eyes and the mouth areas in later layers, our method still achieved state of the art performance. Indeed, the best performance of our proposed method is achieved when keeping 60\% of the image tokens, and even when discarding up to 90\% of the tokens, the results in an accuracy drop of only 0.3\% compared to the baseline, demonstrating its effectiveness even with aggressive token pruning.

Therefore, our main finding is that the FER task does not necessarily need complete facial information, and that specific regions of the face, such as the eyes, mouth, and parts of the cheeks, are sufficient to predict emotional states.
Based on this observation, we propose Sparse Attention to Emotion (SAE), a token pruning approach for the FER task. To the best of our knowledge, this is the first work to explore the application of token pruning as an efficient solution to FER tasks.

Empirical results on the Raf-DB \cite{li2017reliable} dataset show that we generally outperform state of the art methods on computational complexity reduction, while matching, or outperforming them in terms of accuracy. Specifically, our model achieves the best accuracy at a 40\% token discard rate and a competitive performance even after discarding 90\% of tokens. Our contributions are presented as follows:

\begin{itemize}
    \item We propose a token pruning approach for the FER task to reduce the computational cost of ViT models, while limiting their performance drop.
    \item We simplify the complexity of FER by focusing on discriminative facial regions that carry important emotional information.
    \item We validate the robustness and efficiency of our approach, demonstrating that it matches or outperforms previous state of the art methods on the RAF-DB dataset.
\end{itemize}

\section{Related works} 

\subsection{Vision Transformers}

The Vision Transformer (ViT) \cite{dosovitskiy2021an} represents an image as a sequence of patches and uses the attention mechanism to capture the correlations between them. DeiT \cite{touvron2021training} enhances data efficiency by introducing the knowledge distillation token, reducing the need for large-scale datasets required by ViT. Swin Transformer \cite{liu2021swin} introduces hierarchical representation with a shifted window, limiting the attention computation to only local regions, while progressively merging patches. LocalViT \cite{li2023localvit} reintegrates the convolutional local inductive bias into Transformers, improving local feature modeling without sacrificing global context.
    
Despite their effectiveness, these methods display quadratic complexity, as mentioned above, making them costly, which led to the development of compression techniques.  

\subsection{Token reduction}

Multiple works addressed the complexity of Transformers through the reduction of the number of tokens processed. DynamicVit \cite{rao2021dynamicvit} introduces a module that prunes tokens across layers based on their importance. Building on this idea, EViT \cite{liang2022evit} proposes a simple and effective approach that removes inattentive tokens based on the cls-token attention score, significantly reducing the inference cost and without introducing additional parameters. Evo-ViT \cite{xu2022evo} introduces a slow-fast token evaluation strategy, in which informative tokens are selected and processed through all layers, while placeholder tokens evolve through fewer layers. 

Although these methods effectively reduce computational cost, their effectiveness is limited mainly to generic benchmarks such as ImageNet, and their applicability to domain-specific tasks such as facial emotion recognition remains largely unexplored. 

\subsection{Facial Emotion Recognition}

Facial Emotion Recognition (FER) is a task that identifies and classifies human emotions from facial expressions. Many approaches have been proposed to address this task, such as TransFER \cite{Xue_2021_ICCV} which introduces a framework that learns relation-aware facial representations by modeling global interactions between local facial patches, while encouraging the diversity of learned facial features through attention dropping strategies (MAD, MSAD). VTFF \cite{ma2021facial} presents Visual Transformers with Feature Fusion, the method introduces an additional module ASF to combine multi-branch CNN features, and then applies a Transformer to capture global relationships among the resulting tokens, enhancing the robustness to occlusion and pose variation. EAC \cite{zhang2022learn} proposes a method for noisy-label FER by randomly erasing input images and using the flip attention consistency, preventing the model from overfitting to wrong labels. APViT \cite{xue2022vision} introduces two attentive pooling modules, APP is used to select the most attentive patches in CNN features, while APT is used to discard inattentive tokens in ViT, resulting in lower cost and better performance. DAN \cite{wen2023distract} designs facial emotion recognition frameworks consisting of three sub-networks that extract the features, attend to multiple facial regions, fuse features from different heads, and produce the prediction score. POSTER \cite{Zheng_2023_ICCV} introduces a FER framework that combines image features and facial landmark features, using two streams, and by doing a cross fusion attention, it uses a pyramid structure enabling multi-scale handling, addressing the intra-class discrepancy, inter-class similarity, and scale sensitivity. POSTER++ \cite{mao2023postersimplerstrongerfacial} proposes an improved version of POSTER, with the aim of reducing the expensive computational cost. ARBEx \cite{wasi2023arbex} introduces a ViT-based attentive feature extraction framework with a reliability balancing mechanism based on anchors, cross-attention, and confidence estimation. LFNSB \cite{chen2024lightweight} designs a facial recognition network that effectively balances model, complexity, and recognition accuracy, and this by using two key modules, LFN and SN. S2D \cite{chen2024static} presents a framework that transfers knowledge from static facial expression recognition to dynamic facial expression. BTN \cite{Her_2025} designs a framework in FER that learns reliable features by using class batch attention and multi-level attention to reduce noise and overfitting. 

Compared to previous work that processes the entire image, our hypothesis is that not all facial regions are important for the FER task. Only a subset of facial regions is needed for the model to correctly classify an emotion, which is done by applying token pruning to remove unnecessary tokens and retain only the tokens of the most discriminative regions such as the eyes, the mouth, and parts of the cheeks. 

\section{Methodology} 
Here we describe our proposed approach for facial emotion recognition, called Sparse Attention to Emotions. It is built upon token pruning.
\subsection{Self-Attention Mechanism}

Given an input sequence $\mathbf{X}\in\mathbb{R}^{N\times d}$, where $N$ is the number of tokens and $d$ is the embedding dimension, the self-attention mechanism projects the input into queries, keys, values using learnable matrices $\mathbf{W_Q},\mathbf{W_K},\mathbf{W_V} \in \mathbb{R}^{d \times d} $
\begin{equation}
    Q = XW_Q, \quad K=XW_K, \quad V=XW_V
\end{equation}
The attention output is computed as: 
\begin{equation}
\label{eq:attention}
\text{Attention}(Q,K,V) = \text{softmax}\Big(\frac{Q K^\top}{\sqrt{d}}\Big)V.
\end{equation}

The output of the self attention is then passed to a feed forward network (MLP):
\begin{equation}
\text{MLP}(X) = \text{FC}_2(\text{RELU}(\text{FC}_1(X))),
\end{equation}
The total complexity cost of the Transformer layer is given by \cite{pan2021scalable}: 
\begin{equation}
\phi_{\text{BLK}}(N,d) = \phi_{\text{MSA}}(N,d) + \phi_{\text{MLP}}(N,d) = 12 N d^2 + 2 N^2 d,
\end{equation}
highlighting the quadratic complexity with respect to the sequence length $N$.

\subsection{Token Pruning}

Given an input sequence $\mathbf{X}\in\mathbb{R}^{N\times d}$, where $N$ is the number of tokens and $d$ is the embedding dimension. We compute the self-attention matrix following Eq \ref{eq:attention}.

We focus on the first row corresponding to the \texttt{[CLS]} token, each element $a_i$ in this row representing the importance of the $i$-th token:
\begin{equation}
\mathbf{A}_{\text{CLS}} = \text{softmax}\Big(\frac{\mathbf{Q}_{\text{CLS}} \mathbf{K}^\top}{\sqrt{d}}\Big)V
\end{equation}
\begin{equation}
a_i = (\mathbf{A}_{\text{CLS}})_i, \quad i = 1,\dots,N.
\end{equation}
For multi-head attention with $H$ heads, the importance scores are averaged across all heads:
\begin{equation}
\bar{a} = \frac{1}{H} \sum_{h=1}^H a^{(h)}.
\end{equation}
Let $\mathcal{K}$ denote the indices of the top-$k$ tokens that are kept based on their importance scores, and $\mathcal{N}$ denote the indices of the remaining tokens. Inattentive tokens are fused into a single token, and the final token sequence is obtained by concatenating the kept tokens with the fused token. Formally:

\begin{equation}
\mathcal{K} = \text{TopK}(a_1, \dots, a_N; k), \quad
\mathcal{N} = \{1, \dots, N\} \setminus \mathcal{K}
\end{equation}

\begin{equation}
\mathbf{x}_{\text{fused}} = \sum_{i \in \mathcal{N}} a_i \mathbf{x}_i
\end{equation}
\begin{equation}
    \mathbf{X}_{\text{final}} = [\{\mathbf{x}_i\}_{i \in \mathcal{K}}, \mathbf{x}_{\text{fused}}]
\end{equation}

\subsection{Sparse Attention to Emotion}

\begin{figure*}[htpb]
    \centering
    \includegraphics[width=\linewidth]{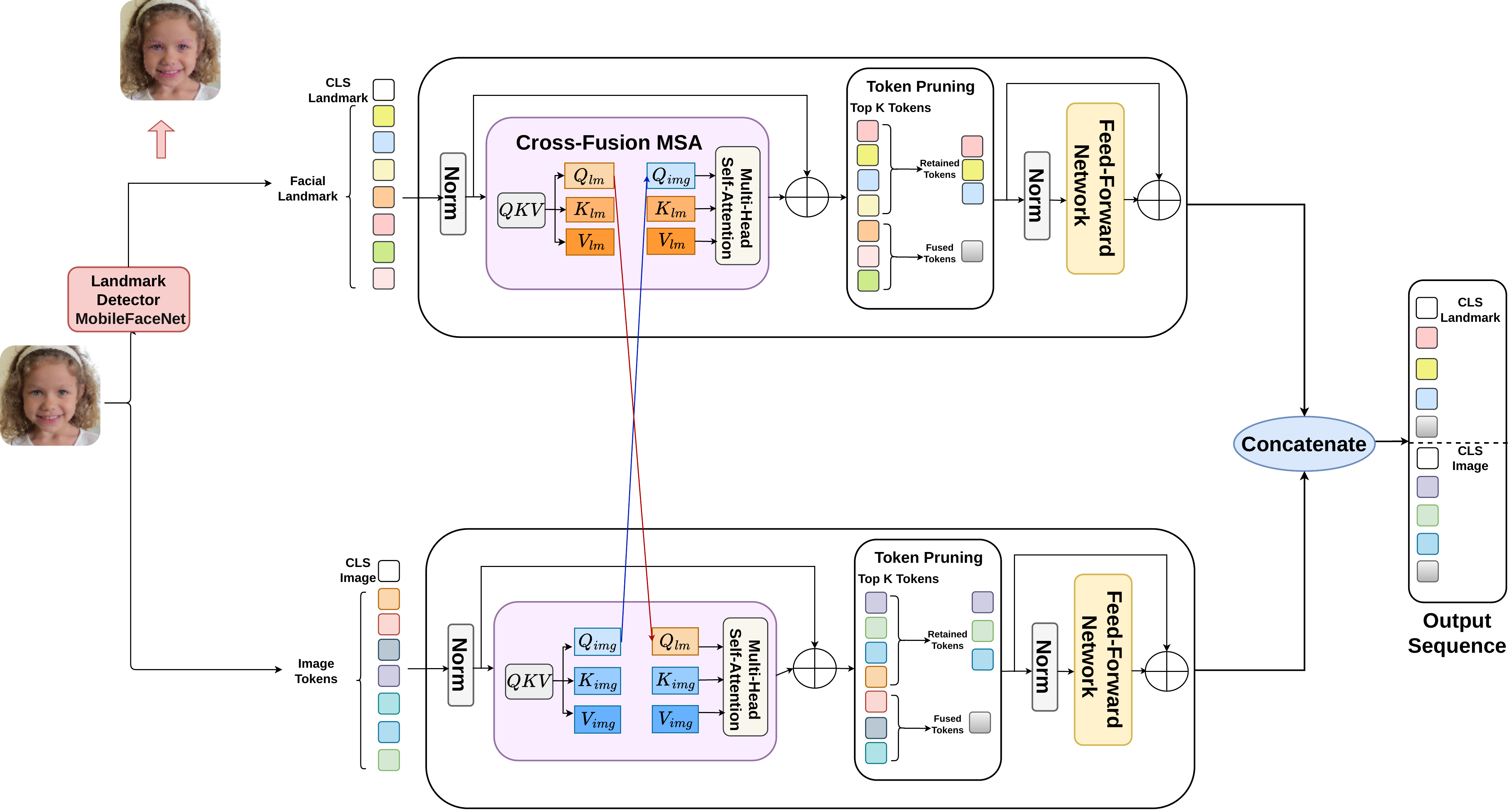}
    \caption{Overview of SAE architecture.}
    \label{fig:SAE_overview}
\end{figure*}

The overview of SAE is shown in Figure \ref{fig:SAE_overview}. Given an input facial image, we extract visual features using a convolutional backbone, while facial landmark features are obtained through a parallel landmark encoder. Let $\mathbf{X} \in \mathbb{R}^{N \times d}$, $ \mathbf{L} \in \mathbb{R}^{N \times d} $ denote the image-based and landmark-based token embeddings, respectively, where $N$ is the number of tokens and $d$ is the embedding dimension. The image and landmark sequences are processed by a cross-fusion multi-head self-attention (MSA) layer. Linear projections are first applied to obtain query, key, and value matrices as follows:

\begin{equation}
\mathbf{Q}_X = \mathbf{X} \mathbf{W}_Q^{(1)}, \quad
\mathbf{K}_X = \mathbf{X} \mathbf{W}_K^{(1)}, \quad
\mathbf{V}_X = \mathbf{X} \mathbf{W}_V^{(1)},
\end{equation}
\begin{equation}
\mathbf{Q}_L = \mathbf{L} \mathbf{W}_Q^{(2)}, \quad
\mathbf{K}_L = \mathbf{L} \mathbf{W}_K^{(2)}, \quad
\mathbf{V}_L = \mathbf{L} \mathbf{W}_V^{(2)}.
\end{equation}

Cross fusion attention is then computed following Eq \ref{eq:attention}:
\begin{equation}
\text{ATTN}_X = \text{Attention}(\mathbf{Q}_L, \mathbf{K}_X, \mathbf{V}_X),
\end{equation}
\begin{equation}
\text{ATTN}_L = \text{Attention}(\mathbf{Q}_X, \mathbf{K}_L, \mathbf{V}_L).
\end{equation}
This cross-attention mechanism enables mutual information exchange between image and landmark representations.
Residual connections are applied to obtain the fused representations:
\begin{equation}
\mathbf{X}' = \text{ATTN}_X + \mathbf{X},
\end{equation}
\begin{equation}
\mathbf{L}' = \text{ATTN}_L + \mathbf{L}.
\end{equation}
The fused sequences are then passed through the token pruning module:
\begin{equation}
\mathbf{X}^{pr} = \mathcal{TP}(\mathbf{X}'), \qquad
\mathbf{L}^{pr} = \mathcal{TP}(\mathbf{L}').
\end{equation}
where $\mathcal{TP}$ represents the token pruning module.
Finally, each pruned sequence is processed by a feed-forward network:
\begin{equation}
\mathbf{X}_{\text{out}} = \text{MLP}(\text{Norm}(\mathbf{X}^{pr})) + \mathbf{X}^{pr},
\end{equation}
\begin{equation}
\mathbf{L}_{\text{out}} = \text{MLP}(\text{Norm}(\mathbf{L}^{pr})) + \mathbf{L}^{pr}.
\end{equation}

\section{Experiments}

\subsection{Datasets and experimental settings}

\noindent\textbf{RAF-DB:} The Real-world Affective Faces Database (RAF-DB) \cite{li2017reliable} is a large-scale facial expression dataset containing real-world facial images. It includes seven emotion categories: surprise, fear, disgust, anger, neutral, sad, and happiness.

\noindent\textbf{Training settings:} The model is trained using the Sharpness Aware Minimization (SAM) optimizer wrapped around the Adam optimizer. The initial learning rate is set to $4\times10^{-5}$, with a batch size of 100, the learning rate is exponentially decayed using ExponentialLR with $\gamma = 0.98$ through the training process for 300 epochs. MobileFaceNet is used as the landmark feature extractor with frozen weights.

\subsection{Experiment results}

Table \ref{tab:comparison} presents the class-wise and mean accuracy of SAE on RAF-DB. For a detailed analysis of SAE performance compared to the state of the art, with a keep rate of 0.9, SAE achieves a mean accuracy of 84.96\%, remaining close to state of the art methods that use full representations. When reducing the keep rate to 0.6, the mean accuracy increases to 85.60\%, confirming that pruning can preserve discriminative tokens across most emotion classes. For keep rate of 0.3, despite discarding 70\% of tokens, SAE maintains a competitive mean accuracy of 85.19\%, with stable performance on dominant classes such as Happy, Neutral, and Sad. Even for a very low keep rate 0.1, where 90\% of tokens are removed, the mean accuracy remains at 85.36\%, a small degradation compared to the other approaches. Across all keep rates, our model shows strong robustness to token reduction, achieving significant low cost while preserving balanced per-class performance.

\begin{table*}[!t]
\centering
\caption{Comparison of SAE with state of the art methods on RAF-DB dataset in terms of class-wise accuracy, and mean Accuracy and keep rate (tokens).}
\label{tab:comparison}
\begin{tabular}{lccccccccccc}
\hline
Method  & \multicolumn{7}{c}{Accuracy of Emotions (\%)} & Mean Acc (\%) & keep rate\\
\cline{2-9}
& Neutral & Happy & Sad & Surprise & Fear & Disgust & Anger \\
\hline
VTFF \cite{ma2021facial} & 87.50 & 94.09 & 87.24 & 85.41 & 64.86 & 68.12 & 85.80 & 81.20&- \\
TransFER \cite{Xue_2021_ICCV}  & 90.15 & 95.95 & 88.70 & 89.06 & 68.92 & 79.37 & 88.89 & 85.86 &- \\
POSTER++ \cite{mao2023postersimplerstrongerfacial} & 92.06 & 97.22 & 92.89 & 90.58 & 68.92 & 71.88 & 88.27 & 85.97  &-\\
POSTER \cite{Zheng_2023_ICCV} & 92.35 & 96.96 & 91.21 & 90.27 & 67.57 & 75.00 & 88.89 & 86.04  &-\\
APViT \cite{xue2022vision} & 92.06 & 97.30 & 88.70 & 93.31 & 72.97 & 73.75 & 86.42 & 86.36  &-\\
BTN \cite{Her_2025} & 92.21 & 97.05 & 92.26 & 91.49 & 72.97 & 76.25 & 88.89 & 87.30  &-\\

\textbf{SAE}& 90.00 & 95.78 & 88.91 & 89.06 & 68.92 & 75.00 & 87.04 & 84.96 & 0.9 \\
\textbf{SAE}  & 92.21 & 96.12 & 90.79 & 89.97 & 63.51 & 72.50 & 85.19 & 85.80 & 0.6 \\
\textbf{SAE}& 90.44 & 96.03 & 92.47 & 87.54 & 70.27 & 75.00 & 84.57 & 85.19 & 0.3 \\
\textbf{SAE}  & 91.18 & 96.29 & 89.96 & 89.67 & 70.27 & 71.88 & 88.27 & 85.36 & 0.1 \\

\hline
\end{tabular}
\end{table*}

\begin{table}[!ht]
\caption{Comparison of SAE with state of the art approaches on RAF-DB in terms of overall accuracy (\%) and keep rate (tokens).}
\label{tab:SOTA_summary}
\centering
\begin{tabular}{lcc}
\hline
\textbf{Methods} & \textbf{Keep rate} & \textbf{overall accuracy} \\

\hline
TransFER \cite{Xue_2021_ICCV}       & -   & 90.91  \\
EfficientFace \cite{zhao2021robust}& -   & 88.36  \\
EAC \cite{zhang2022learn}          & -   & 90.35  \\
APViT \cite{xue2022vision}         & -   & 91.98 \\
DAN \cite{wen2023distract}         & -   & 89.70  \\
POSTER \cite{Zheng_2023_ICCV}      & -   & 92.05  \\
POSTER++ \cite{mao2023postersimplerstrongerfacial} & - & 92.21  \\
ARBEx \cite{wasi2023arbex}         & -   & 92.47  \\
LFNSB \cite{chen2024lightweight}   & -   & 91.07  \\
S2D \cite{chen2024static}          & -   & 92.21  \\
S2D* \cite{chen2024static}         & -   & 92.57  \\
BTN \cite{Her_2025}                 & -   & 92.54 \\
\textbf{SAE}                        & 0.9 & 90.51 \\
\textbf{SAE}                        & 0.6 & 91.17 \\
\textbf{SAE}                        & 0.3 & 91.00 \\
\textbf{SAE}                        & 0.1 & 91.13  \\
\hline
\end{tabular}
\end{table}

Table \ref{tab:SOTA_summary} reports the overall accuracy of SAE compared to state of the art methods on RAF-DB, while focusing on the trade-off between performance and computational complexity. For a high keep rate 0.9, SAE reaches an accuracy of 90.51\%, which is already comparable to other approaches. When the keep rate is reduced to 0.6, the accuracy increases to 91.13\%, remaining competitive with recent transformer-based methods. Even with more aggressive pruning for a keep rate of 0.3 and 0.1, SAE shows strong robustness, the accuracy remains stable. These results clearly indicate that SAE can reduce computational complexity by up to 90\% while causing only a negligible performance loss often below 1\% demonstrating an efficient accuracy complexity trade-off compared to existing approaches.

\section{Conclusion}
In this paper, we proposed Sparse Attention to Emotion, an efficient approach that explores token pruning in facial emotion recognition to reduce computational complexity without compromising performance. SAE integrates a pyramid cross-fusion architecture that exploits complementary information from landmark and image streams, and uses multi-scale feature representation to capture facial emotions at different resolutions. Experiments on common benchmark datasets showed that SAE matches or outperforms state of the art results while retaining only 10\% of the image tokens, making it well suited for edge deployment and resource constrained applications.


\bibliographystyle{IEEEbib}
\bibliography{strings,refs}

\end{document}